\documentclass[10pt,twocolumn,letterpaper]{article}

\usepackage{wacv}              

\usepackage{graphicx}
\usepackage{amsmath}
\usepackage{amssymb}
\usepackage{booktabs}

\usepackage[accsupp]{axessibility}  

\usepackage{graphicx}
\usepackage{times}
\usepackage{epsfig}
\usepackage{graphicx}
\usepackage{amssymb}
\usepackage{xcolor}
\usepackage{caption}
\usepackage{tabularx}
\usepackage{rotating}
\usepackage{verbatim}
\usepackage{xcolor,colortbl}
\usepackage{etoolbox}
\usepackage[normalem]{ulem}
\usepackage{booktabs}
\usepackage{multirow}
\usepackage{lipsum}
\usepackage{stmaryrd}
\usepackage{stackengine}
\usepackage{makecell}
\usepackage{pifont}
\usepackage{cancel}
\usepackage{adjustbox}
\usepackage{dblfloatfix}
\usepackage{amsmath}
\usepackage{booktabs}

\definecolor{LightGrey}{rgb}{.9,.9,.9}
\definecolor{White}{rgb}{1.,0.,1.}
\definecolor{first}{rgb}{.8,.0,.0}
\definecolor{second}{rgb}{.0,.6,.0}
\definecolor{third}{rgb}{.0,.0,.8}
\newcolumntype{g}{>{\columncolor{White}}c}

\definecolor{car}{rgb}{0.39215686, 0.58823529, 0.96078431}
\definecolor{bicycle}{rgb}{0.39215686, 0.90196078, 0.96078431}
\definecolor{motorcycle}{rgb}{0.11764706, 0.23529412, 0.58823529}
\definecolor{truck}{rgb}{0.31372549, 0.11764706, 0.70588235}
\definecolor{other-vehicle}{rgb}{0.39215686, 0.31372549, 0.98039216}
\definecolor{person}{rgb}{1.        , 0.11764706, 0.11764706}
\definecolor{bicyclist}{rgb}{1.        , 0.15686275, 0.78431373}
\definecolor{motorcyclist}{rgb}{0.58823529, 0.11764706, 0.35294118}
\definecolor{road}{rgb}{1.        , 0.        , 1.        }
\definecolor{parking}{rgb}{1.        , 0.58823529, 1.        }
\definecolor{sidewalk}{rgb}{0.29411765, 0.        , 0.29411765}
\definecolor{other-ground}{rgb}{0.68627451, 0.        , 0.29411765}
\definecolor{building}{rgb}{1.        , 0.78431373, 0.        }
\definecolor{fence}{rgb}{1.        , 0.47058824, 0.19607843}
\definecolor{vegetation}{rgb}{0.        , 0.68627451, 0.        }
\definecolor{trunk}{rgb}{0.52941176, 0.23529412, 0.        }
\definecolor{terrain}{rgb}{0.58823529, 0.94117647, 0.31372549}
\definecolor{pole}{rgb}{1.        , 0.94117647, 0.58823529}
\definecolor{traffic-sign}{rgb}{1.        , 0.        , 0.    }

\newcommand\crule[3][black]{\textcolor{#1}{\rule{#2}{#3}}}
\definecolor{nvcolor}{RGB}{119,185,0}
\definecolor{roadcolor}{RGB}{234,51,246}
\definecolor{sidewalkcolor}{RGB}{68,8,72}
\definecolor{parkingcolor}{RGB}{241,156,249}
\definecolor{othergroundcolor}{RGB}{160,32,76}
\definecolor{buildingcolor}{RGB}{246,202,69}
\definecolor{carcolor}{RGB}{111,149,238}
\definecolor{truckcolor}{RGB}{74,32,172}
\definecolor{bicyclecolor}{RGB}{136,227,242}
\definecolor{motorcyclecolor}{RGB}{37,59,146}
\definecolor{othervehiclecolor}{RGB}{96,81,242}
\definecolor{vegetationcolor}{RGB}{79, 173, 50}
\definecolor{trunkcolor}{RGB}{126, 65, 22}
\definecolor{terraincolor}{RGB}{171, 238, 105}
\definecolor{personcolor}{RGB}{234, 60, 49}
\definecolor{bicyclistcolor}{RGB}{234, 66, 195}
\definecolor{motorcyclistcolor}{RGB}{138, 42, 90}
\definecolor{fencecolor}{RGB}{238, 128, 69}
\definecolor{polecolor}{RGB}{252, 241, 161}
\definecolor{trafficsigncolor}{RGB}{233, 51, 35}
\definecolor{other-struct.color}{RGB}{255, 150, 0}
\definecolor{other-objectcolor}{RGB}{50, 255, 255}
\definecolor{lane-markingcolor}{RGB}{150, 255, 170}
\definecolor{color1}{RGB}{176, 36, 24}
\definecolor{color2}{RGB}{0, 176, 80}
\definecolor{color3}{RGB}{0, 0, 200}
\definecolor{colorofteaser}{RGB}{176, 36, 24}

\usepackage{colortbl}
\definecolor{mygray}{gray}{0.9}

\makeatletter
\newcommand{\car@semkitfreq}{3.92}
\newcommand{\bicycle@semkitfreq}{0.03}
\newcommand{\motorcycle@semkitfreq}{0.03}
\newcommand{\truck@semkitfreq}{0.16}
\newcommand{\othervehicle@semkitfreq}{0.20}
\newcommand{\person@semkitfreq}{0.07}
\newcommand{\bicyclist@semkitfreq}{0.07}
\newcommand{\motorcyclist@semkitfreq}{0.05}
\newcommand{\road@semkitfreq}{15.30}  %
\newcommand{\parking@semkitfreq}{1.12}
\newcommand{\sidewalk@semkitfreq}{11.13}  %
\newcommand{\otherground@semkitfreq}{0.56}
\newcommand{\building@semkitfreq}{14.1}  %
\newcommand{\fence@semkitfreq}{3.90}
\newcommand{\vegetation@semkitfreq}{39.3}  %
\newcommand{\trunk@semkitfreq}{0.51}
\newcommand{\terrain@semkitfreq}{9.17} %
\newcommand{\pole@semkitfreq}{0.29}
\newcommand{\trafficsign@semkitfreq}{0.08}
\newcommand{\semkitfreq}[1]{{\csname #1@semkitfreq\endcsname}}

\newcommand{\ceiling@nyufreq}{1.37}
\newcommand{\floor@nyufreq}{17.58}
\newcommand{\wall@nyufreq}{15.26}
\newcommand{\window@nyufreq}{1.99}
\newcommand{\chair@nyufreq}{3.01}
\newcommand{\bed@nyufreq}{7.08}
\newcommand{\sofa@nyufreq}{4.70}
\newcommand{\table@nyufreq}{4.31}
\newcommand{\tvs@nyufreq}{0.47}
\newcommand{\furniture@nyufreq}{30.04}
\newcommand{\objects@nyufreq}{14.19}
\newcommand{\nyufreq}[1]{{\csname #1@nyufreq\endcsname}}

\usepackage[pagebackref,breaklinks,colorlinks]{hyperref}

\usepackage[capitalize]{cleveref}
\crefname{section}{Sec.}{Secs.}
\Crefname{section}{Section}{Sections}
\Crefname{table}{Table}{Tables}
\crefname{table}{Tab.}{Tabs.}

\def\wacvPaperID{68} 
\def\confName{WACV}
\def\confYear{2025}

\begin{document}

\title{DepthSSC: Monocular 3D Semantic Scene Completion via Depth-Spatial Alignment and Voxel Adaptation}

\author{Jiawei Yao$^1$ \quad Jusheng Zhang$^2$ \quad Xiaochao Pan$^3$ \quad Tong Wu$^1$ \quad Canran Xiao$^4$\thanks{Corresponding author.}\\
$^1$ University of Washington
$^2$ Sun Yat-sen University\\
$^3$ Taiyuan University of Technology
$^4$ Central South University\\
{\tt\small jwyao@uw.edu, xiaocanran@csu.edu.cn} \\
}

\maketitle

\begin{abstract}
The task of 3D semantic scene completion using monocular cameras is gaining significant attention in the field of autonomous driving. This task aims to predict the occupancy status and semantic labels of each voxel in a 3D scene from partial image inputs. Despite numerous existing methods, many face challenges such as inaccurately predicting object shapes and misclassifying object boundaries. To address these issues, we propose DepthSSC, an advanced method for semantic scene completion using only monocular cameras. DepthSSC integrates the Spatial Transformation Graph Fusion (ST-GF) module with Geometric-Aware Voxelization (GAV), enabling dynamic adjustment of voxel resolution to accommodate the geometric complexity of 3D space. This ensures precise alignment between spatial and depth information, effectively mitigating issues such as object boundary distortion and incorrect depth perception found in previous methods. Evaluations on the SemanticKITTI and SSCBench-KITTI-360 dataset demonstrate that DepthSSC not only captures intricate 3D structural details effectively but also achieves state-of-the-art performance.

\end{abstract}    
\section{Introduction}
\label{sec:intro}

\begin{figure}[t]
    \centering
    \includegraphics[width=\columnwidth]{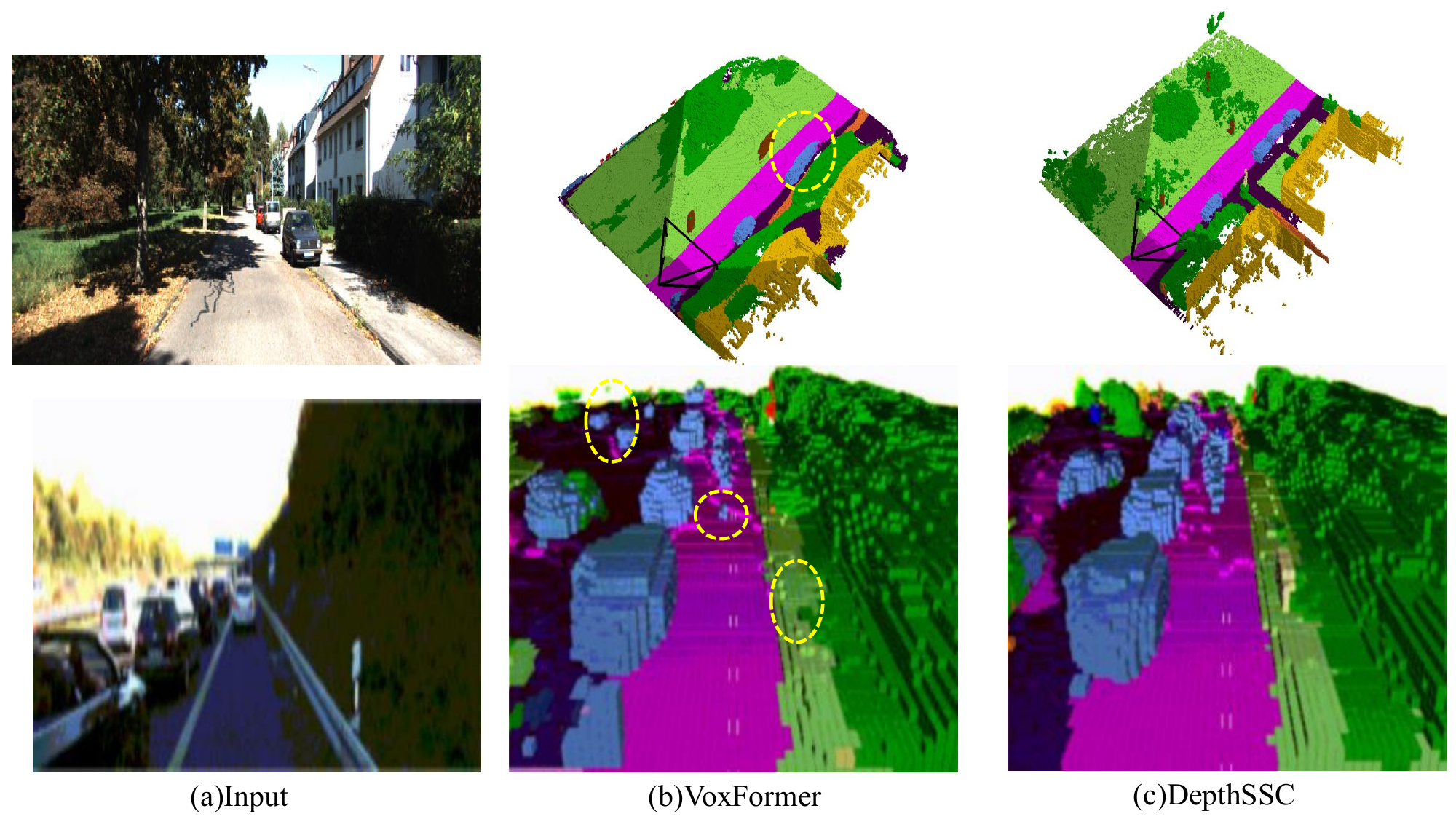}
    \caption{\textbf{Goal of Our Approach.} Demonstrates DepthSSC's superiority in handling complex 3D environments for semantic scene completion. Contrasted with VoxFormer, DepthSSC excels in accurately predicting occupancy grids for both nearby and distant objects, overcoming geometric complexity. The highlighted areas showcase VoxFormer's challenges with misrecognition and non-recognition. As illustrated in the first row, VoxFormer fails to adequately utilize depth relationships to distinguish between different vehicles. In the second row, under complex lighting conditions, key features of objects at varying distances are lost, preventing VoxFormer from effectively inferring the 3D structure of the scene. Our DepthSSC addresses this issue by taking into account the geometric complexity of 3D spaces and allowing for dynamic adjustment of voxel resolution. }
   \vspace{-0.7cm}
    \label{fig:intro}
\end{figure}

3D Semantic Scene Completion (SSC)~\cite{song2017semantic} is crucial for autonomous driving, predicting voxel occupancy in 3D scenes from partial inputs. S3cnet\cite{cheng2021s3cnet} and Scpnet\cite{xia2023scpnet}, which leverage LiDAR-generated point clouds, are examples among various approaches that have been developed for 3D semantic scene completion. Recent approaches have shifted towards vision-centric methods, such as SSCNet~\cite{song2017semantic}, MonoScene~\cite{cao2022monoscene}, NDC-Scene~\cite{yao2023ndc}, and OccDepth~\cite{miao2023occdepth}, which elevate 2D image attributes into 3D spaces using 3D convolutional networks. TPVFormer~\cite{huang2023tri} and VoxFormer~\cite{li2023voxformer} further enhance scene modeling with Transformer architectures. Additionally, Occ3D~\cite{tian2023occ3d} and OccFormer~\cite{zhang2023occformer} feature coarse-to-fine designs and class-specific mask prediction strategies. H2GFormer~\cite{wang2024h2gformer} introduces a horizontal-to-global voxel transformer for improved semantic feature fusion, while HASSC~\cite{wang2024not} employs a hardness-aware design and self-distillation strategy to enhance SSC accuracy without extra inference cost.

However, previous visual approaches~\cite{cao2022monoscene, li2023voxformer, yao2023ndc} have faced significant challenges when reconstructing accurate 3D scenes from monocular camera data. The limitations of these cameras, such as the lack of stereoscopic depth perception and restricted field of view, often lead to spatial distortions and deformations. These issues result in inaccuracies in object shapes and misclassifications of boundaries, ultimately affecting critical autonomous driving tasks like object detection, path planning, and obstacle avoidance. For instance, as illustrated in Figure~\ref{fig:intro}
, VoxFormer~\cite{li2023voxformer} struggles to accurately predict occupancy grids in complex environments, frequently misrecognizing or failing to recognize objects due to poor depth relationship utilization and loss of key features under varying lighting conditions. Accurate 3D scene completion requires precise spatial alignment to correctly identify object boundaries and maintain relationships between objects.

Furthermore, while Transformer-based architectures like TPVFormer~\cite{huang2023tri} and VoxFormer~\cite{li2023voxformer} have shown promise in adaptively extracting voxel features, they still fall short in addressing the geometric complexity of real-world 3D environments. These models often use uniform voxel resolutions across different regions, failing to capture the intricate details and structures accurately~\cite{xiao2024debsdf}. This oversight can result in the loss of crucial information in complex areas and waste computational resources on simpler regions.

In light of these challenges, our work introduces DepthSSC, a novel method designed to address the limitations of monocular SSC. DepthSSC is inspired by how we use spatial and depth cues to infer the 3D structure of scenes in driving scenarios. For example, when viewing a row of parked cars along the street from a single image, we can estimate their distances and spatial arrangement by recognizing the relative sizes and positions of the cars, as well as their alignment with road markings and surrounding objects. Similarly, DepthSSC leverages spatial transformation and geometric awareness to ensure precise spatial-depth alignment and adaptive voxel resolution. By integrating the Spatially-Transformed Graph Fusion (ST-GF) module and Geometrically-aware Voxelization, our approach aligns spatial data with depth cues and dynamically adjusts voxel resolutions to accurately capture the complexity of different scene regions. Extensive experiments on the SemanticKITTI~\cite{behley2019semantickitti} dataset demonstrate superior performance, and ablation studies confirm the effectiveness of our components. Our key contributions are:

\begin{itemize}
    \item We propose DepthSSC, a new method that integrates spatial transformation with geometric awareness to address the issues of inaccurate depth perception and spatial distortions in monocular 3D SSC. DepthSSC achieves state-of-the-art results with 15.22 mIoU and 44.89 IoU on the SemanticKITTI benchmark (hidden test set), surpassing the latest approaches.
    \item We introduce the Spatially-Transformed Graph Fusion module, which facilitates the spatial transformation and feature fusion from voxels to graph structures, ensuring precise alignment of spatial and depth information within the same spatial framework.
    \item We propose Geometrically-aware Voxelization, a method that considers the geometric complexity of 3D space, allowing for dynamic adjustments of voxel resolution to adapt to various regions.
\end{itemize}

\section{Related Work}
\label{sec:formatting}

\paragraph{3D Semantic Scene Completion.} Understanding 3D scenes is essential for applications like robotic perception, autonomous driving, and digital twins. SSCNet~\cite{song2017semantic} showed that semantic segmentation and scene completion are interconnected, leading to the concept of Semantic Scene Completion (SSC). Current SSC methods fall into three categories based on input data: depth maps~\cite{zhang2019cascaded, wang2019forknet, li2019depth}, depth maps with RGB images~\cite{li2020anisotropic, li2019rgbd}, point clouds~\cite{cheng2021s3cnet, rist2021semantic}, and RGB images~\cite{cao2022monoscene, li2023voxformer, yao2023ndc}. Depth map-based methods predict voxel occupancy and semantic labels from a single depth map but miss the color and texture richness of RGB images. Models like TS3D~\cite{garbade2019two} use dual-stream networks to combine RGB and depth data, enhancing SSC performance. Point cloud-based SSC excels in large-scale outdoor scenes but lacks a standard feature extraction approach. Recently, RGB image-based methods have gained popularity due to their cost-effectiveness and potential for real-world applications. Efficient methods like spatial grouped convolution (SGC)~\cite{zhang2018efficient} reduce computational costs by focusing on effective voxel regions without sacrificing accuracy.

\vspace{-0.5cm}

\paragraph{Monocular 3D Semantic Scene Completion.} Completing 3D scenes using only RGB images is challenging. Initial works like MonoScene~\cite{cao2022monoscene} and OccDepth~\cite{miao2023occdepth} project 2D features into 3D spaces using depth perception. BEVFormer~\cite{li2022bevformer}, SurroundOcc~\cite{wei2023surroundocc}, and TPVFormer~\cite{huang2023tri} aggregate 2D features within 3D space using query-based mechanisms. VoxFormer~\cite{li2023voxformer} expands sparse voxel queries based on depth into dense 3D structures. H2GFormer~\cite{wang2024h2gformer} enhances feature fusion focusing on voxel distinctions in outdoor scenarios. HASSC~\cite{wang2024not} uses hardness-aware design and self-distillation for SSC accuracy in difficult regions. Symphonies~\cite{jiang2024symphonize} integrates instance queries for better context in scenes with occlusion and perspective errors. Li et al.~\cite{li2023lode} extend Eikonal-based approaches to non-watertight LiDAR point clouds for large-scale scene completion, significantly improving performance on datasets like SemanticKITTI. Bi-SSC~\cite{xue2024bi} enhances semantic correlation with spatial sensory fusion across single and stereo views. MonoOcc~\cite{zheng2024monoocc} improves prediction with auxiliary semantic loss and cross-attention modules. Despite these advancements, monocular methods often face difficulties in accurately capturing complex 3D scene structures and details. Challenges such as insufficient depth perception, occlusion handling, and uniform voxel resolution limit their effectiveness.

\begin{figure*}[t]
    \centering
    \includegraphics[width=0.7\textwidth]{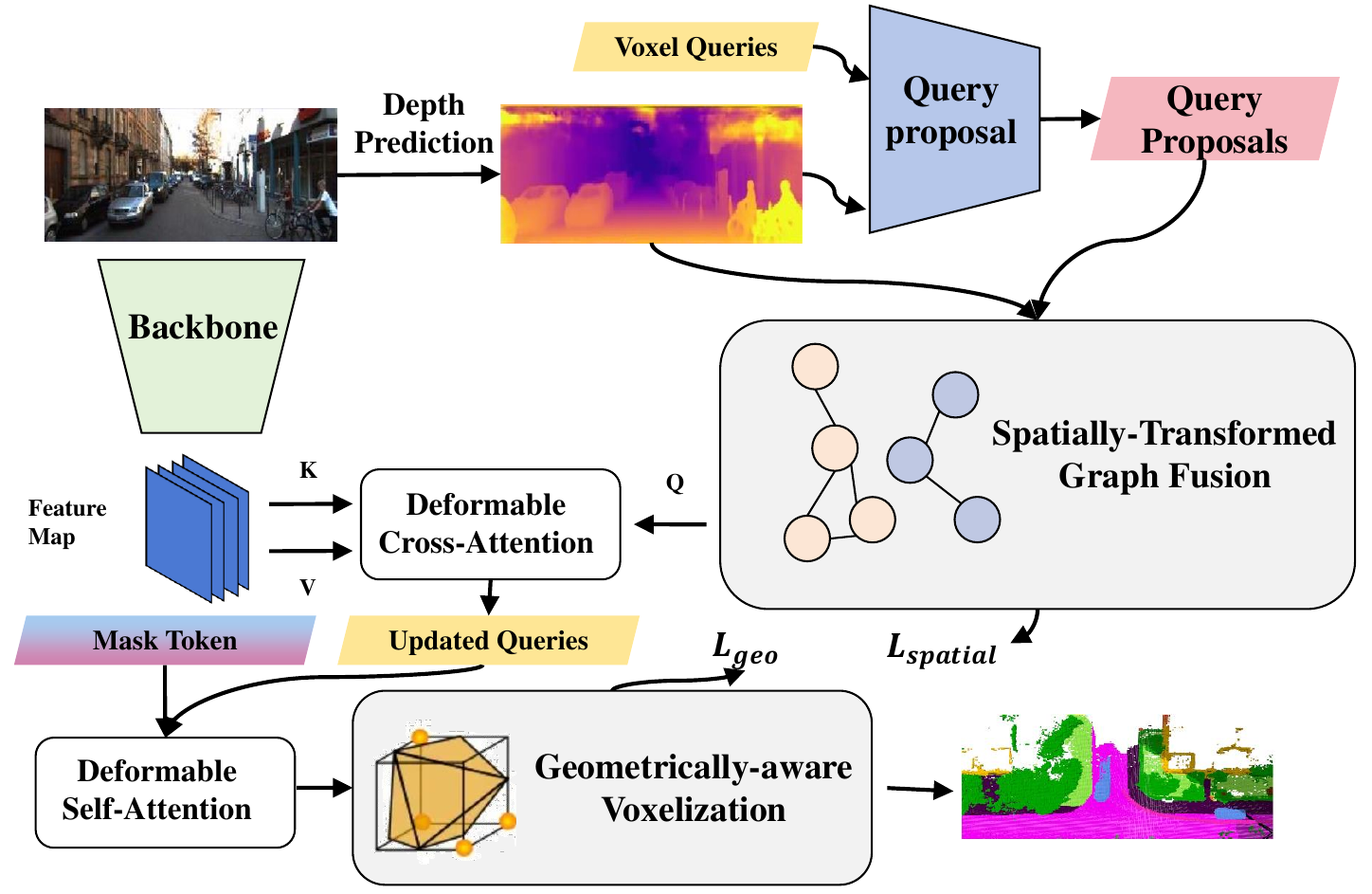}
            \caption{\textbf{Pipeline of the DepthSSC.} The process begins with a backbone extracting 2D features from input images, followed by the projection of these 2D features into the 3D voxel domain using deformable self-attention and deformable cross-attention mechanisms. These mechanisms integrate spatial and geometric information. The Spatially-Transformed Graph Fusion (ST-GF) module corrects spatial misalignments by predicting a 3D affine transformation matrix for voxel queries, forming a graph structure that enables feature fusion via Graph Convolutional Networks (GCNs). Geometrically-aware Voxelization (GAV) adapts voxel resolution dynamically based on the geometric complexity of regions, enhancing the model’s ability to represent intricate structures and improve overall 3D scene completion.}

   \vspace{-0.4cm}
    \label{fig:overview}
\end{figure*}
\section{Method}
\label{sec:method}

In this part, we first introduce the baseline model VoxFormer~\cite{li2023voxformer} in Section~\ref{sec:preliminary}. Then, we will present two core modules, Spatially-Transformed Graph Fusion and Geometrically-aware Voxelization, in Sections~\ref{sec:Spatially-Transformed Graph Fusion} and~\ref{Geometrically-aware Voxelization}, respectively. The architecture of our proposed method is shown in Figure~\ref{fig:overview}.

\subsection{Preliminary}
\label{sec:preliminary}
VoxFormer leverages the power of Transformers to enhance 3D Semantic Scene Completion (SSC) by using voxel queries, which are 3D grid-shaped learnable parameters that capture and project 2D image features into the 3D domain using camera projection matrices. The process is divided into two stages: generating class-agnostic query proposals and class-specific segmentation, with a ResNet-50 backbone for 2D feature extraction.

From an input RGB image \( I_t \), 2D features \( F_{2D_t} \in \mathbb{R}^{b \times c \times d} \) are extracted using a convolutional neural network backbone, where \( b \times c \) is the spatial resolution and \( d \) is the feature dimension.

Voxel queries \( Q \in \mathbb{R}^{h \times w \times z \times d} \) are 3D-grid-shaped parameters that map 2D features to the 3D volume, with \( h \times w \times z \) set to a lower resolution to reduce computational load. A subset of voxel queries \( Q_p = \text{Reshape}(Q[M_{\text{out}}]) \) is selected based on predicted occupancy from depth information, forming \( Q_p \in \mathbb{R}^{N_p \times d} \), where \( M_{\text{out}} \) is the corrected occupancy map.

The Deformable Cross-Attention (DCA) mechanism allows flexible attention over a local neighborhood in the 2D feature space using calculated offsets:
\begin{equation}
\texttt{DCA}(\mathbf{q}_p, \mathbf{F}^{2D}) = \frac{1}{|\mathcal{V}_{t}|} \sum_{t \in \mathcal{V}_{t}} \texttt{DA}(\mathbf{q}_p, \mathcal{P}(\mathbf{p}, t), \mathbf{F}_t^{2D}),
\end{equation}
where \( P(p, t) \) maps 3D points to the 2D plane of image \( t \).

The Deformable Self-Attention (DSA) mechanism refines voxel features by enabling interactions within the 3D space:
\begin{equation}
\texttt{DSA}(\mathbf{F}^{3D}, \mathbf{F}^{3D}) = \texttt{DA}(\mathbf{f}, \mathbf{p}, \mathbf{F}^{3D}),
\end{equation}
where \( f \) is a mask token or updated query proposal at location \( p \).

The final output \( Y_t \in \mathbb{R}^{H \times W \times Z \times (M+1)} \) represents the semantic segmentation map, where \( H \times W \times Z \) is the output resolution and \( M+1 \) indicates \( M \) semantic classes plus one void class.

\subsection{Spatially-Transformed Graph Fusion}
\label{sec:Spatially-Transformed Graph Fusion}
In 3D SSC, accurately aligning voxel queries with depth maps is critical for generating a coherent 3D representation. Due to independent predictions from neural networks, minor inconsistencies in the voxel features and depth information can result in spatial misalignments, causing distortions in the reconstructed 3D scene. These misalignments can negatively impact the final prediction accuracy, particularly in regions with complex geometries or near object boundaries.

\begin{figure*}[t]
    \centering
    \includegraphics[width=0.9\textwidth]{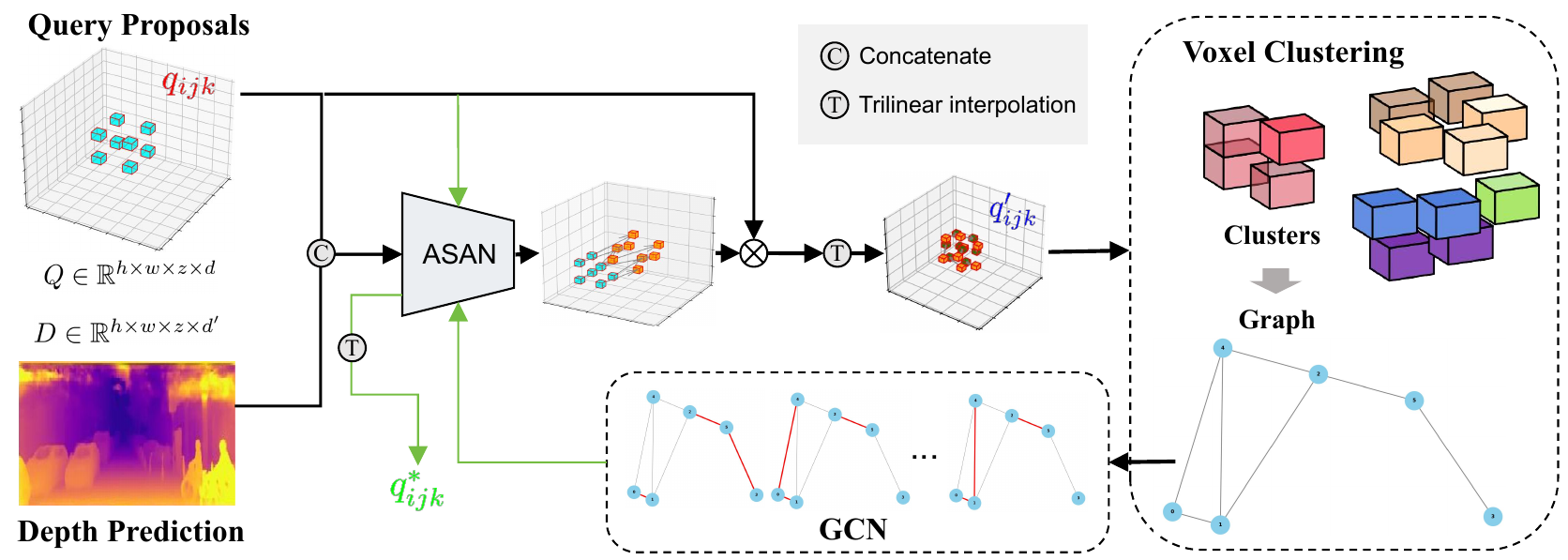}
            \caption{\textbf{Workflow of the Spatially-Transformed Graph Fusion (ST-GF) Module.} The ST-GF module corrects spatial misalignments by predicting a 3D affine transformation matrix \( \Theta_{ijk} \). Voxels are then clustered based on spatial similarity and processed through a Graph Convolutional Network (GCN) to aggregate features. Finally, an inverse affine transformation and trilinear interpolation are applied, ensuring accurate spatial alignment and enhancing 3D scene coherence.}           
            
   \vspace{-0.5cm}
    \label{fig:STGF}
\end{figure*}

To address these spatial alignment issues, we propose the \textit{Spatially-Transformed Graph Fusion (ST-GF)} module, as shown in Figure~\ref{fig:STGF}. The ST-GF module is designed with three primary objectives: (1) correcting geometric distortions by predicting a 3D affine transformation matrix \( \Theta_{ijk} \), which allows flexible adjustments in voxel position, including rotation, scaling, and translation; (2) capturing local geometric structures by clustering voxels into spatially similar regions, enabling better modeling of complex shapes; and (3) capture spatial dependencies between voxels, especially in geometrically intricate areas, ensuring coherent 3D scene representation.

\vspace{-0.3cm}
\paragraph{Voxel-to-Node Mapping.} 
To better align voxel features and depth information, we first concatenate the voxel queries \( Q \in \mathbb{R}^{h \times w \times z \times d} \) and the depth predictions \( D \in \mathbb{R}^{h \times w \times z \times d'} \) along the feature dimension, producing \( Q' \in \mathbb{R}^{h \times w \times z \times (d + d')} \). The \textit{Adaptive Spatial Adjustment Network (ASAN)} then predicts a 3D affine transformation matrix \( \Theta_{ijk} \) for each voxel \( q_{ijk} \in Q' \), allowing for the flexible adjustment of voxel positions to better align with their true spatial locations. 

ASAN is a neural network that predicts this affine transformation matrix \( \Theta_{ijk} \), which consists of rotation, scaling, and translation components. Specifically, the transformation matrix \( \Theta_{ijk} \in \mathbb{R}^{4 \times 4} \) is defined as:
\begin{equation}
\label{eq:theta}
\Theta_{ijk} = 
\begin{bmatrix}
R_{ijk} & T_{ijk} \\
0 & 1
\end{bmatrix}
\end{equation}
where \( R_{ijk} \in \mathbb{R}^{3 \times 3} \) is the rotation and scaling matrix, and \( T_{ijk} \in \mathbb{R}^{3} \) is the translation vector.

ASAN predicts the rotation angles \( \theta, \phi, \psi \) and scaling factors \( s_x, s_y, s_z \) for each voxel. The rotation matrix \( R_{ijk} \) is parameterized as:
\begin{equation}
\label{eq:R}
R_{ijk}(\theta, \phi, \psi) = 
\begin{bmatrix}
c_\theta c_\phi & c_\theta s_\phi s_\psi - s_\theta c_\psi & c_\theta s_\phi c_\psi + s_\theta s_\psi \\
s_\theta c_\phi & s_\theta s_\phi s_\psi + c_\theta c_\psi & s_\theta s_\phi c_\psi - c_\theta s_\psi \\
-s_\phi & c_\phi s_\psi & c_\phi c_\psi
\end{bmatrix}
\end{equation}
where \( c_\theta \) and \( s_\theta \) represent \( \cos(\theta) \) and \( \sin(\theta) \), respectively, for the angles \( \theta, \phi, \psi \).

The scaling factors \( s_x, s_y, s_z \) are predicted alongside the rotation and incorporated into the rotation matrix. The scaling matrix \( S(s_x, s_y, s_z) \) is defined as:
\begin{equation}
\label{eq:S}
S(s_x, s_y, s_z) = 
\begin{bmatrix}
s_x & 0 & 0 \\
0 & s_y & 0 \\
0 & 0 & s_z
\end{bmatrix}
\end{equation}
Thus, the combined rotation and scaling matrix \( R_{ijk} \) becomes:
\begin{equation}
R_{ijk} = R(\theta, \phi, \psi) \cdot S(s_x, s_y, s_z)
\end{equation}
In addition to rotation and scaling, ASAN predicts the translation vector \( T_{ijk} = [t_x, t_y, t_z]^T \), which adjusts the voxel’s position in 3D space. We substitute equation \ref{eq:S}, equation \ref{eq:R}, and \( T_{ijk} \) into equation \ref{eq:theta}, and obtain:
\begin{equation}
\Theta_{ijk} = 
\begin{bmatrix}
\begin{smallmatrix}
c_\theta c_\phi s_x & (c_\theta s_\phi s_\psi - s_\theta c_\psi) s_y & (c_\theta s_\phi c_\psi + s_\theta s_\psi) s_z & t_x \\
s_\theta c_\phi s_x & (s_\theta s_\phi s_\psi + c_\theta c_\psi) s_y & (s_\theta s_\phi c_\psi - c_\theta s_\psi) s_z & t_y \\
-s_\phi s_x & c_\phi s_\psi s_y & c_\phi c_\psi s_z & t_z \\
0 & 0 & 0 & 1
\end{smallmatrix}
\end{bmatrix}
\end{equation}

Once the transformation is applied, the grid generator creates a sampling grid to map the original space \( P_{out} \) to the new space \( P_{in} \):
\begin{equation}
P_{in} = \Theta \times P_{out}
\end{equation}

After the affine transformation, the voxels are no longer strictly aligned to the regular 3D grid. Trilinear interpolation \cite{csebfalvi2019beyond} is applied to adjust the voxel positions, ensuring the continuity of voxels in the new grid positions.
\begin{equation}
q_{ijk}' = \text{Trilinear}(q_{ijk}, P_{in})
\end{equation}
where, $q_{ijk}'$ represents the transformed voxel position. The result is a set of transformed voxels \( q_{ijk}' \), with improved spatial alignment.

\vspace{-0.3cm}
\paragraph{Voxel Clustering and Graph Structure Construction.} After transforming the voxel positions, we cluster these voxels based on spatial similarity to construct the graph structure.

Given the transformed voxels \( q_{ijk}' \), we perform the following steps. For each voxel \( q_{ijk}' \), compute the Euclidean distance to its neighboring voxels:
\begin{equation}
d(q_{ijk}', q_{i'j'k'}') = \sqrt{(x_i - x_{i'})^2 + (y_j - y_{j'})^2 + (z_k - z_{k'})^2}
\end{equation}

Use a soft clustering approach to assign each voxel \( q_{ijk}' \) to clusters. Let \( \pi_{ijk}^m \) represent the probability of voxel \( q_{ijk}' \) belonging to cluster \( m \):
\begin{equation}
\pi_{ijk}^m = \frac{\exp(-\alpha d(q_{ijk}', \mu_m))}{\sum_{n=1}^M \exp(-\alpha d(q_{ijk}', \mu_n))}
\end{equation}
where \( \alpha \) is a scaling factor.
 
Calculate the center \( \mu_m \) of each cluster \( C_m \) as a weighted average:
\begin{equation}
\mu_m = \frac{\sum_{ijk} \pi_{ijk}^m q_{ijk}'}{\sum_{ijk} \pi_{ijk}^m}
\end{equation}

Using the clusters and their centers, we construct the graph. Each cluster \( C_m \) is represented as a node \( v_m \). Create an edge \( e_{mn} \) between nodes \( v_m \) and \( v_n \) if:
\begin{equation}
e_{mn} = 
\begin{cases}
1 & \text{if} \ d(\mu_m, \mu_n) < \epsilon \\
0 & \text{otherwise}
\end{cases}
\end{equation}

where \( d(\mu_m, \mu_n) = \sqrt{(\mu_{m,x} - \mu_{n,x})^2 + (\mu_{m,y} - \mu_{n,y})^2 + (\mu_{m,z} - \mu_{n,z})^2} \) and \( \epsilon \) is a predefined threshold.

Construct the adjacency matrix \( A \), where \( A_{mn} = e_{mn} \). The graph \( G = (V, E) \) with nodes \( V \) and edges \( E \) captures the spatial relationships among voxel clusters, facilitating subsequent feature fusion and optimization.

\vspace{-0.3cm}
\paragraph{Graph Fusion.}
After constructing the graph, we apply Graph Convolutional Networks (GCNs)~\cite{chen2020simple} to aggregate features from neighboring nodes and refine voxel proposals. For each node \( v_m \), graph convolution updates its features based on its neighbors \( \mathcal{N}(m) \):
\begin{equation}
F_m^{(l+1)} = \sigma \left( \sum_{n \in \mathcal{N}(m)} A_{mn} W_g^{(l)} F_n^{(l)} + b_g^{(l)} \right),
\end{equation}
where \( A \) is the adjacency matrix and \( \sigma \) is the ReLU activation.

The refined node features \( F_m^{(L)} \) are mapped back to voxel proposals:
\begin{equation}
Q_{proposal} = \sum_{m=1}^M \pi_{ijk}^m F_m^{(L)},
\end{equation}
where \( \pi_{ijk}^m \) is the soft cluster assignment. 

Finally, we processed the voxel proposals through inverse affine transformation and trilinear interpolation to obtain refined voxel positions that are spatially coherent and aligned with the depth map:
\begin{equation}
q_{ijk}^* = \text{Trilinear}(\text{ASAN}^{-1}(Q_{proposal}, \Theta), P_{out}),
\end{equation}

\subsection{Geometrically-aware Voxelization}
\label{Geometrically-aware Voxelization}
Voxformer's deformable self-attention targets local regions for richer visual features but struggles with fixed voxel resolution in geometrically complex areas. This limitation affects precise geometry modeling. To address this, we introduce Geometrically-aware Voxelization (GAV), which dynamically adjusts voxel resolution based on geometric complexity. Higher resolution captures intricate details in complex regions, while lower resolution conserves resources in simpler areas.

\vspace{-0.3cm}
\paragraph{Geometric Complexity Assessment.} For a given three-dimensional voxel feature \( S \), we use the Marching Cubes~\cite{newman2006survey} algorithm to obtain the surface \( M \). For each voxel \( V_i \), we compute its geometric complexity \( C(V_i) \), defined as the number of intersections of the voxel with the surface \( M \):
\begin{equation}
C(V_i) = \left| \{p \in V_i \cap M\} \right|,
\end{equation}
where \( \left| \cdot \right| \) denotes the cardinality of the set.

For each voxel \( V_i \), we map its geometric complexity \( C(V_i) \) to a continuous range representing its resolution:
\begin{equation}
R(V_i) = f(C(V_i)),
\end{equation}
where \( f \) is a sigmoid activation function.

\begin{figure*}[t]
    \centering
    
    \includegraphics[width=0.8\textwidth]{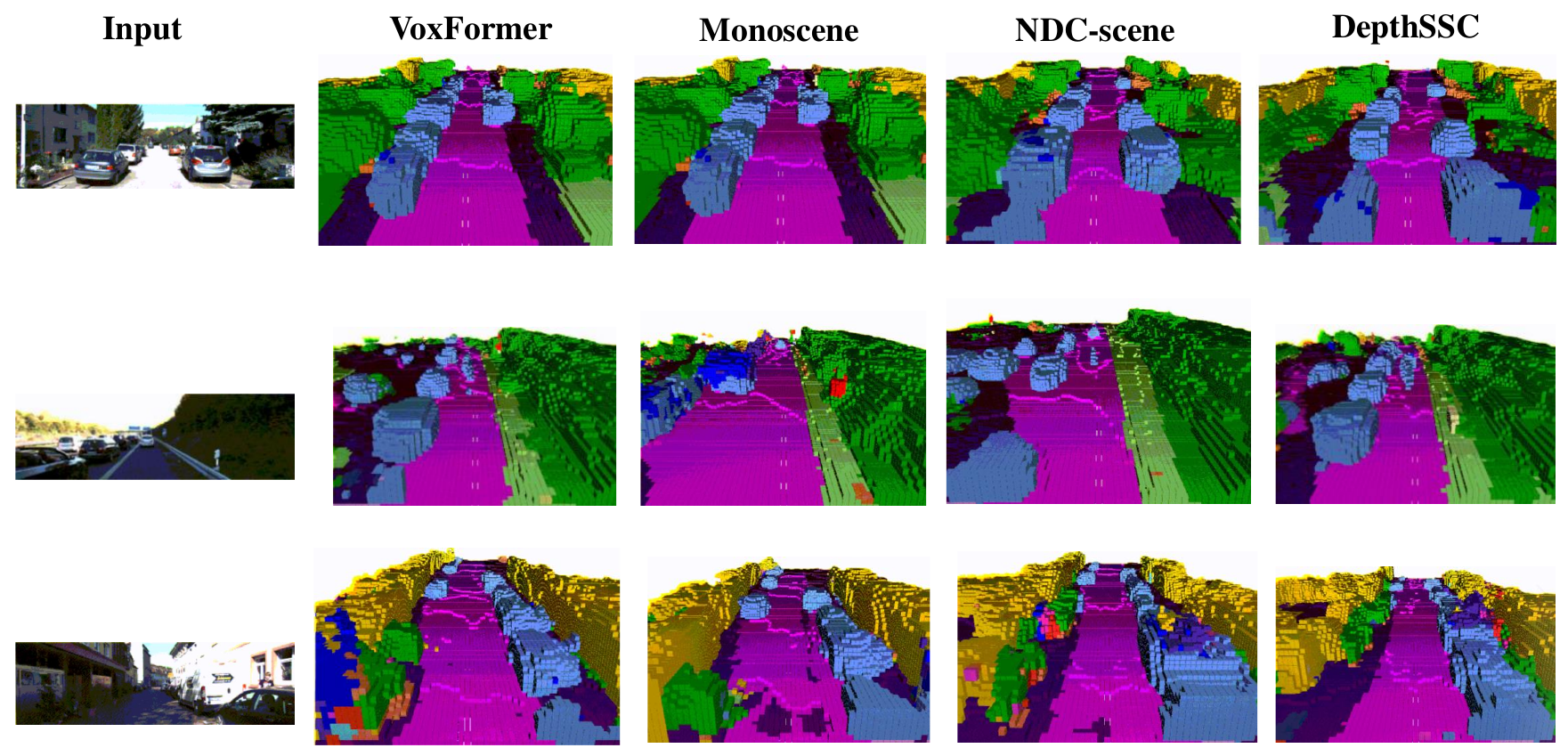}
\vspace{-0.3cm}
    \caption{\textbf{Visualization results} against the state-of-the-art monocular SSC methods on SemanticKITTI~\cite{behley2019semantickitti} (val set). }
    
 \vspace{-0.5cm}
    \label{fig:vis}
\end{figure*}

\vspace{-0.4cm}

\paragraph{Resolution-Adaptive Deformable Attention.} When performing deformable self-attention operations, we consider the dynamic resolution \( R(V_i) \) of voxels. The resolution directly affects the position and quantity of query points in deformable self-attention, with higher resolution corresponding to finer query points. For each voxel \( V_i \), its position in three-dimensional space can be represented as \( p = (x, y, z) \). We adjust these positions based on the resolution \( R(V_i) \), allowing voxels with higher complexity to have a higher query density.

First, we calculate the offset of query points, \( \Delta p = W_p f \), where \( W_p \) is a weight matrix and \( f \) represents the feature map or an updated query proposal. Next, we update the query point positions, \( p' = p + \delta R(V_i) + \Delta p \), where \( \delta \) is a constant determining the influence of resolution \( R(V_i) \) on query point positions.

With the new query positions \( p' \), we compute the updated queries \( Q' = W_q p' \) and calculate the deformable attention weights:
\begin{equation}
   DA(f, p', F_{3D}) = \operatorname{softmax}\left(\frac{Q' K^T}{\sqrt{d_k}}\right) V. 
\end{equation}

The refined voxel features \( \hat{F}_{3D} \) using deformable self-attention are obtained as \( \hat{F}_{3D} = DA(f, p', F_{3D}) \).

\subsection{Training Loss}

For the first-stage training, we focus on spatial continuity and occupancy prediction. We use binary cross-entropy loss for occupancy prediction at a lower spatial resolution. To ensure spatial continuity during the fusion process, we define a loss by comparing distances between neighboring nodes:
\begin{align}
&L_{spatial}=\frac1M\sum_{j=1}^M\|d(G_{orig,j},G_{orig,j+1}) \notag \\
&-d(G_{transformed,j},G_{transformed,j+1})\|.
\end{align}
where \( d() \) is the distance function between nodes, and \( M \) is the total number of nodes.

In the second stage, the model is trained for multi-class semantic voxel grid prediction using weighted cross-entropy loss and Scene-class Affinity Loss. Additionally, to prevent geometric shape distortion or detail loss due to Geometrically-aware Voxelization, we introduce Geometric Preservation Loss \( L_{geo} \) based on the Hausdorff distance\cite{huttenlocher1993comparing} between the original 3D data points \( P \) and the voxelized data points \( Q \):

\begin{equation}
    L_{geo}=d_H(P,Q).
\end{equation}

\section{Experiments}

In this section, we verify the proposed DepthSSC on SemanticKITTI and SSCBench-KITTI-360 dataset  and compare DepthSSC against previous approaches in Section~\ref{results}. We provide abundant ablation studies in Section~\ref{ablation} for a comprehensive understanding of the proposed method. The experimental setup is provided in the supplementary materials, where the dataset, metrics, and baselines are described in detail.


\subsection{Main Results}

\label{results}
We conducted experiments to evaluate the performance of the DepthSSC on the SemanticKITTI~\cite{behley2019semantickitti} and SSCBench-KITTI-360~\cite{li2023sscbench}, and compared its results against other state-of-the-art camera-based models.

\begin{table*}[h]
		\scriptsize
		\captionsetup{font=footnotesize}
		\setlength{\tabcolsep}{0.004\linewidth}
		\newcommand{\classfreq}[1]{{~\tiny(\semkitfreq{#1}\%)}}  %
		\centering
        {\tiny
		\begin{tabular}{l|c|c|c c c c c c c c c c c c c c c c c c c|c}
			\toprule
			& & SC & \multicolumn{20}{c}{SSC} \\
			Method
			& SSC Input
			& {IoU}
			& \rotatebox{90}{\textcolor{road}{$\blacksquare$} road\classfreq{road}} 
			& \rotatebox{90}{\textcolor{sidewalk}{$\blacksquare$} sidewalk\classfreq{sidewalk}}
			& \rotatebox{90}{\textcolor{parking}{$\blacksquare$} parking\classfreq{parking}} 
			& \rotatebox{90}{\textcolor{other-ground}{$\blacksquare$} other-grnd\classfreq{otherground}} 
			& \rotatebox{90}{\textcolor{building}{$\blacksquare$} building\classfreq{building}} 
			& \rotatebox{90}{\textcolor{car}{$\blacksquare$} car\classfreq{car}} 
			& \rotatebox{90}{\textcolor{truck}{$\blacksquare$} truck\classfreq{truck}} 
			& \rotatebox{90}{\textcolor{bicycle}{$\blacksquare$} bicycle\classfreq{bicycle}} 
			& \rotatebox{90}{\textcolor{motorcycle}{$\blacksquare$} motorcycle\classfreq{motorcycle}} 
			& \rotatebox{90}{\textcolor{other-vehicle}{$\blacksquare$} other-veh.\classfreq{othervehicle}} 
			& \rotatebox{90}{\textcolor{vegetation}{$\blacksquare$} vegetation\classfreq{vegetation}} 
			& \rotatebox{90}{\textcolor{trunk}{$\blacksquare$} trunk\classfreq{trunk}} 
			& \rotatebox{90}{\textcolor{terrain}{$\blacksquare$} terrain\classfreq{ &terrain}} 
			& \rotatebox{90}{\textcolor{person}{$\blacksquare$} person\classfreq{person}} 
			& \rotatebox{90}{\textcolor{bicyclist}{$\blacksquare$} bicyclist\classfreq{bicyclist}} 
			& \rotatebox{90}{\textcolor{motorcyclist}{$\blacksquare$} motorcyclist.\classfreq{motorcyclist}} 
			& \rotatebox{90}{\textcolor{fence}{$\blacksquare$} fence\classfreq{fence}} 
			& \rotatebox{90}{\textcolor{pole}{$\blacksquare$} pole\classfreq{pole}} 
			& \rotatebox{90}{\textcolor{traffic-sign}{$\blacksquare$} traf.-sign\classfreq{trafficsign}} 
			& mIoU\\
			\midrule
	
            LMSCNet$^\text{rgb}$~\cite{roldao2020lmscnet} & Occ & 28.61 & 40.68 & 18.22 & 4.38 & 0.00 & 10.31 & 18.33 & 0.00 & 0.00 & 0.00 & 0.00 & 13.66 & 0.02 & 20.54 & 0.00 & 0.00 & 0.00 & 1.21 & 0.00 & 0.00 & 6.70 \\
\hline

            AICNet$^\text{rgb}$~\cite{li2020anisotropic} & RGB \& Depth & 29.59 & 43.55 & 20.55 & 11.97 & 0.07 & 12.94 & 14.71 & 4.53 & 0.00 & 0.00 & 0.00 & 15.37 & 2.90 & 28.71 & 0.00 & 0.00 & 0.00 & 2.52 & 0.06 & 0.00 & 8.31  \\
\hline

            JS3C-Net$^\text{rgb}$~\cite{yan2021sparse}& Pts & 34.00 & 47.30 & 21.70 & 19.90 & \textcolor{blue}{\textbf{2.80}} & 12.70 & 20.10 & 0.80 & 0.00 & 0.00 & 4.10 & 14.20 & 3.10 & 12.40 & 0.00 & 0.20 &  \textcolor{blue}{\textbf{0.20}} &  8.70 &  1.90 &  0.30 &  8.97
            \\

\hline
            
            MonoScene~\cite{cao2022monoscene} & RGB &  37.12 & 57.47 & 27.05 & 15.72 & 0.87 & 14.24 & 23.55 & 7.83 & 0.20 & 0.77 & 3.59 & 18.12 & 2.57 & 30.76 & 1.79 & 1.03 & 0.00 & 6.39 & 4.11 & 2.48 & 11.50 \\

            TPVFormer~\cite{li2020anisotropic} & RGB & 35.61	& 56.50	& 25.87 & 20.60 & 0.85 & 13.88	& 23.81	& 8.08	& 0.36	& 0.05	& 4.35	& 16.92	& 2.26	& 30.38	& 0.51	& 0.89	& 0.00	& 5.94	& 3.14	& 1.52	& 11.36  \\

            Voxformer~\cite{li2023voxformer} & RGB & 44.02 & 54.76 & 26.35 & 15.50 & 0.70 & 17.65 & 25.79 & 5.63 & 0.59 & 0.51 & 3.77 & 24.39 & 5.08 & 29.96 & 1.78 & 3.32 & 0.00 & 7.64 & 7.11 & 4.18 & 12.35 \\

            NDC-Scene~\cite{yao2023ndc} & RGB & 37.24 & 59.20 & 28.24 & \textcolor{blue}{\textbf{21.42}} & 1.67 & 14.94 & 26.26 &  14.75 & 1.67 & 2.37 & 7.73 & 19.09 & 3.51 & 31.04 & \textcolor{blue}{\textbf{3.60}} & 2.74 & 
            0.00 & 6.65 & 4.53 & 2.73 & 12.70\\ 

            HASSC~\cite{wang2024not} & RGB & 44.82 & 57.05 & 28.25 & 15.90 & 1.04 & 19.05 & \textcolor{blue}{\textbf{27.23}} &  9.91 & 0.92 & 0.86 & 5.61 & 25.48 & \textcolor{blue}{\textbf{6.15}} & 32.94 & 2.80 & 4.71 & 
            0.00 & 6.58 & 7.68 & 4.05 & 13.48\\

            DepthSSC (ours) & RGB & \textcolor{blue}{\textbf{45.97}} & \textcolor{blue}{\textbf{59.48}} & \textcolor{blue}{\textbf{29.03}} & 18.81 & 0.99 & \textcolor{blue}{\textbf{19.27}} & 26.22 &  \textcolor{blue}{\textbf{15.44}} & \textcolor{blue}{\textbf{1.83}} & \textcolor{blue}{\textbf{2.39}} & \textcolor{blue}{\textbf{7.82}} & \textcolor{blue}{\textbf{26.92}} & 5.73 & \textcolor{blue}{\textbf{33.47}} & 2.58 & \textcolor{blue}{\textbf{6.27}} & 
            0.00 & \textcolor{blue}{\textbf{8.83}} & \textcolor{blue}{\textbf{7.74}} & \textcolor{blue}{\textbf{4.21}} & \textcolor{blue}{\textbf{14.59}}\\ 
            
			\bottomrule
		\end{tabular}\\
  }
  \vspace{-0.2cm}
	\caption{\textbf{Quantitative comparison} against RGB-inferred baselines and the state-of-the-art monocular SSC method on SemanticKITTI~\cite{behley2019semantickitti} (val set). The best results compared to the corresponding
 baselines are marked in \textcolor{blue}{blue}.}
	\label{tab:SemanticKITTI_val}
\vspace{-0.7cm}
\end{table*}
\vspace{-0.2cm}

\paragraph{SemanticKITTI} As shown in Table~\ref{tab:SemanticKITTI_test}, DepthSSC achieved an IoU of 44.89\% and mIoU of 15.22\%. Compared to other models, DepthSSC outperformed all others in terms of IoU, highlighting its superior capability in semantic scene completion tasks. On the SemanticKITTI validation set, presented in Table~\ref{tab:SemanticKITTI_val}, DepthSSC maintained consistent performance with an IoU of 45.97\% and mIoU of 14.59\%. It again surpassed other models in IoU, showcasing its robustness across different data splits. When examining specific classes, DepthSSC demonstrated exceptional performance in recognizing 'building', 'car', 'vegetation', and 'fence', with scores significantly higher than most other models. However, there are areas for improvement in classes such as 'person' and 'motorcyclist'. The high variability and irregular shapes of these classes present a challenge for the ST-GF module, which aims to align and fuse features accurately. The transformation parameters and clustering mechanisms may not fully account for the non-rigid structures of 'person' and 'motorcyclist' classes, resulting in less precise alignments with their complex geometries.

\begin{table*}
		\scriptsize
		\captionsetup{font=footnotesize}
		\setlength{\tabcolsep}{0.004\linewidth}
		\newcommand{\classfreq}[1]{{~\tiny(\semkitfreq{#1}\%)}}  %
		\centering
        
        {\tiny
		\begin{tabular}{l|c|c|c c c c c c c c c c c c c c c c c c c|c}
			\toprule
			& & SC & \multicolumn{20}{c}{SSC} \\
			Method
			& SSC Input
			& {IoU}
			& \rotatebox{90}{\textcolor{road}{$\blacksquare$} road\classfreq{road}} 
			& \rotatebox{90}{\textcolor{sidewalk}{$\blacksquare$} sidewalk\classfreq{sidewalk}}
			& \rotatebox{90}{\textcolor{parking}{$\blacksquare$} parking\classfreq{parking}} 
			& \rotatebox{90}{\textcolor{other-ground}{$\blacksquare$} other-grnd\classfreq{otherground}} 
			& \rotatebox{90}{\textcolor{building}{$\blacksquare$} building\classfreq{building}} 
			& \rotatebox{90}{\textcolor{car}{$\blacksquare$} car\classfreq{car}} 
			& \rotatebox{90}{\textcolor{truck}{$\blacksquare$} truck\classfreq{truck}} 
			& \rotatebox{90}{\textcolor{bicycle}{$\blacksquare$} bicycle\classfreq{bicycle}} 
			& \rotatebox{90}{\textcolor{motorcycle}{$\blacksquare$} motorcycle\classfreq{motorcycle}} 
			& \rotatebox{90}{\textcolor{other-vehicle}{$\blacksquare$} other-veh.\classfreq{othervehicle}} 
			& \rotatebox{90}{\textcolor{vegetation}{$\blacksquare$} vegetation\classfreq{vegetation}} 
			& \rotatebox{90}{\textcolor{trunk}{$\blacksquare$} trunk\classfreq{trunk}} 
			& \rotatebox{90}{\textcolor{terrain}{$\blacksquare$} terrain\classfreq{ &terrain}} 
			& \rotatebox{90}{\textcolor{person}{$\blacksquare$} person\classfreq{person}} 
			& \rotatebox{90}{\textcolor{bicyclist}{$\blacksquare$} bicyclist\classfreq{bicyclist}} 
			& \rotatebox{90}{\textcolor{motorcyclist}{$\blacksquare$} motorcyclist.\classfreq{motorcyclist}} 
			& \rotatebox{90}{\textcolor{fence}{$\blacksquare$} fence\classfreq{fence}} 
			& \rotatebox{90}{\textcolor{pole}{$\blacksquare$} pole\classfreq{pole}} 
			& \rotatebox{90}{\textcolor{traffic-sign}{$\blacksquare$} traf.-sign\classfreq{trafficsign}} 
			& mIoU\\
			\midrule
	
            LMSCNet$^\text{rgb}$~\cite{roldao2020lmscnet} & Occ & 31.38 & 46.70 & 19.50 & 13.50 & 3.10 & 10.30 & 14.30 & 0.30 & 0.00 & 0.00 & 0.00 & 10.80 & 0.00 & 10.40 & 0.00 & 0.00 & 0.00 & 5.40 & 0.00 & 0.00 & 7.07 \\
\hline

            AICNet$^\text{rgb}$~\cite{li2020anisotropic} & RGB \& Depth & 23.93	& 39.30	& 18.30 & 19.80 & 1.60 & 9.60	& 15.30	& 0.70	& 0.00	& 0.00	& 0.00	& 9.60	& 1.90	& 13.50	& 0.00	& 0.00	& 0.00	& 5.00	& 0.10	& 0.00	& 7.09  \\
\hline

            JS3C-Net$^\text{rgb}$~\cite{yan2021sparse}& Pts & 34.00 & 47.30 & 21.70 & 19.90 & 2.80 & 12.70 & 20.10 & 0.80 & 0.00 & 0.00 & 4.10 & 14.20 & 3.10 & 12.40 & 0.00 & 0.20 &  0.20 &  8.70 &  1.90 &  0.30 &  8.97  \\

            \hline
			MonoScene~\cite{cao2022monoscene} & RGB &  34.16 & 54.70 & 27.10 & 24.80 & 5.70 & 14.40 & 18.80 & 3.30 & 0.50 & 0.70 & 4.40 & 14.90 & 2.40 & 19.50 & 1.00 & 1.40 & 0.40 & 11.10 & 3.30 & 2.10 & 11.08 \\

            TPVFormer~\cite{li2020anisotropic} & RGB & 34.25	& 55.10	& 27.20 & 27.40 & 6.50 & 14.80	& 19.20	& 3.70	& 1.00	& 0.50	& 2.30	& 13.90	& 2.60	& 20.40	& 1.10	& 2.40	& 0.30	& 11.00	& 2.90	& 1.50	& 11.26  \\

            Voxformer~\cite{li2023voxformer} & RGB & 42.95 & 53.90 & 25.30 & 21.10 & 5.60 & 19.80 & 20.80 & 3.50 & 1.00 & 0.70 & 3.70 & 22.40 & 7.50 & 21.30 & 1.40 & 2.60 & 0.20 & 11.10 & 5.10 & 4.90 & 12.20 \\

            NDC-Scene~\cite{yao2023ndc} & RGB & 36.19 & 58.12 & 28.05 & 25.31 & 6.53 & 14.90 & 19.13 &  \textcolor{blue}{\textbf{4.77}} & 1.93 & 2.07 & 6.69 & 17.94 & 3.49 & \textcolor{blue}{\textbf{25.01}} & 3.44 & 2.77 & 
            1.64 & 12.85 & 4.43 & 2.96 & 12.58\\ 

            Symphonies~\cite{jiang2024symphonize} & RGB & 42.19 & 58.40 & 29.30 & \textcolor{blue}{\textbf{26.90}} & \textcolor{blue}{\textbf{11.70}} & 24.70 & 23.60 &  3.20 & 3.60 & 2.60 & 5.60 & 24.20 & \textcolor{blue}{\textbf{10.00}} & 23.10 & 3.20 & 1.90 & 
            \textcolor{blue}{\textbf{2.00}} & \textcolor{blue}{\textbf{16.10}} & 7.70 & 8.00 & 15.04\\ 

                HASSC\cite{wang2024not} & RGB & 43.40 & - & - & - & - & - & - &  - & - & - & - & - & - & - & - & - & 
            - &- & - & - & 13.34\\ 

            DepthSSC (ours) & RGB & \textcolor{blue}{\textbf{44.89}} & \textcolor{blue}{\textbf{59.07}} & \textcolor{blue}{\textbf{29.33}} & 26.43 & 9.16 & \textcolor{blue}{\textbf{25.43}} & \textcolor{blue}{\textbf{24.06}} &  4.51 & \textcolor{blue}{\textbf{3.97}} & \textcolor{blue}{\textbf{2.90}} & \textcolor{blue}{\textbf{6.82}} & \textcolor{blue}{\textbf{25.36}} & 9.41 & 23.72 & \textcolor{blue}{\textbf{3.88}} & \textcolor{blue}{\textbf{2.89}} & 
            0.68 & 15.37 & \textcolor{blue}{\textbf{7.98}} & \textcolor{blue}{\textbf{8.22}} & \textcolor{blue}{\textbf{15.22}}\\ 
            
			\bottomrule
		\end{tabular}\\
    }
    \vspace{-0.3cm}
        \caption{\textbf{Quantitative comparison} against RGB-inferred baselines and the state-of-the-art monocular SSC method on SemanticKITTI~\cite{behley2019semantickitti} (hidden test set).The best results compared to the corresponding
 baselines are marked in \textcolor{blue}{blue}.}

	\label{tab:SemanticKITTI_test}
 \vspace{-0.5cm}
\end{table*}
\begin{table}[t]\centering

\scriptsize

\setlength{\tabcolsep}{0.004\linewidth}
		\newcommand{\classfreq}[1]{{~\tiny(\semkitfreq{#1}\%)}}  %
\resizebox{\columnwidth}{!}{%

\begin{tabular}{l|ccc|ccc|cccc}\toprule

\textbf{Methods}  &\multicolumn{3}{c|}{\textbf{VoxFormer-S}} &\multicolumn{3}{c|}{\textbf{MonoScene}} &\multicolumn{3}{c}{\textbf{DepthSSC}} \\\midrule

\textbf{Range (m)}  & 12.8m&25.6m&   51.2m&  12.8m& 25.6m&  51.2m&  12.8m& 25.6m& 51.2m   \\\midrule

\textbf{IoU (\%)}  
   & 55.45&46.36&    38.76& 54.65&44.70 &  37.87&\textcolor{blue}{\textbf{59.37}}&\textcolor{blue}{\textbf{49.47}}&\textcolor{blue}{\textbf{40.85}}  \\

\textbf{Precision (\%)}& 66.10&61.34&    58.52& 65.88& 59.96& 56.73 &\textcolor{blue}{\textbf{67.42}}&\textcolor{blue}{\textbf{63.75}}&\textcolor{blue}{\textbf{60.69}} \\

\textbf{Recall (\%)}&77.48& 65.48&53.44&76.24&63.72&	53.26&\textcolor{blue}{\textbf{77.55}}&\textcolor{blue}{\textbf{65.96}}&\textcolor{blue}{\textbf{55.86}}  \\ \midrule

\textbf{mIoU}  & 18.17&	15.40&	11.91&	20.29&	16.18&	12.31&	\textcolor{blue}{\textbf{20.52}} &\textcolor{blue}{\textbf{17.44}}& \textcolor{blue}{\textbf{14.28}}   \\

\crule[carcolor]{0.13cm}{0.13cm} \textbf{car} (2.85\%) & 29.41&	25.08&	17.84&	30.83&	26.35&	19.34&\textcolor{blue}{\textbf{33.72}}&\textcolor{blue}{\textbf{28.20}}&\textcolor{blue}{\textbf{21.90}}   \\

\crule[bicyclecolor]{0.13cm}{0.13cm} \textbf{bicycle} (0.01\%)  & \textcolor{blue}{\textbf{2.73}}&	1.73&	1.16	&1.94	&0.83&	0.43&	1.43&\textcolor{blue}{\textbf{2.85}}&\textcolor{blue}{\textbf{2.36}}&\\

\crule[motorcyclecolor]{0.13cm}{0.13cm} \textbf{motorcycle} (0.01\%)& 1.97&	1.47&	0.89&	3.25&	1.30&	0.58&\textcolor{blue}{\textbf{5.19}}&\textcolor{blue}{\textbf{2.15}}&\textcolor{blue}{\textbf{4.30 }}
\\

\crule[truckcolor]{0.13cm}{0.13cm} \textbf{truck} (0.16\%)  & 6.08&	6.63&	4.56&	14.83&	12.18&	8.02&\textcolor{blue}{\textbf{16.03}}&\textcolor{blue}{\textbf{16.03}}&\textcolor{blue}{\textbf{11.51 }}\\

\crule[othervehiclecolor]{0.13cm}{0.13cm} \textbf{other-veh.} (5.75\%) & 3.71&	3.56 &	2.06 &	6.08&	4.30& 	2.03&\textcolor{blue}{\textbf{6.73}}&\textcolor{blue}{\textbf{5.13}}&\textcolor{blue}{\textbf{4.56}}  \\

\crule[personcolor]{0.13cm}{0.13cm} \textbf{person} (0.02\%)  &2.86&\textcolor{blue}{\textbf{2.20}} &	1.63 &	2.06 &	1.26 &	0.86&\textcolor{blue}{\textbf{3.71}} &	1.16&\textcolor{blue}{\textbf{2.92}} &  \\

\crule[roadcolor]{0.13cm}{0.13cm} \textbf{road} (14.98\%) & 66.10 & 58.58 & 47.01 & 68.60  &59.93  &48.35  &\textcolor{blue}{\textbf{72.28}}&\textcolor{blue}{\textbf{63.47}} & \textcolor{blue}{\textbf{50.88}}   
  \\
\crule[parkingcolor]{0.13cm}{0.13cm} \textbf{parking} (2.31\%) & 18.44& 13.52 &9.67&\textcolor{blue}{\textbf{24.32}}&\textcolor{blue}{\textbf{16.40}}&11.38 &	21.72&	15.10&\textcolor{blue}{\textbf{12.89}}& 
\\
\crule[sidewalkcolor]{0.13cm}{0.13cm} \textbf{sidewalk} (6.43\%) &38.00 &	33.63 &	27.21& 	44.43 &	36.05 &	28.13&\textcolor{blue}{\textbf{48.79}}&\textcolor{blue}{\textbf{42.41}}&\textcolor{blue}{\textbf{30.27 }} \\

\crule[othergroundcolor]{0.13cm}{0.13cm} \textbf{other-grnd}(2.05\%) & 4.49 &4.04 &2.89 &\textcolor{blue}{\textbf{5.76}}&\textcolor{blue}{\textbf{4.82}} &\textcolor{blue}{\textbf{3.32}}&	3.87 &	4.77 &	2.49 \\

\crule[buildingcolor]{0.13cm}{0.13cm} \textbf{building} (15.67\%) & 41.12 &	38.24 	&31.18 &	45.40 &	\textcolor{blue}{\textbf{40.60}} &	32.89	&\textcolor{blue}{\textbf{46.03}} &	38.65 &	\textcolor{blue}{\textbf{37.33}} \\

\crule[fencecolor]{0.13cm}{0.13cm} \textbf{fence} (0.96\%) &8.99 	&\textcolor{blue}{\textbf{7.43}}& 	4.97 &	9.79 &	5.91 &	3.53	&\textcolor{blue}{\textbf{10.87}} &	6.81 &\textcolor{blue}{\textbf{5.22}} \\

\crule[vegetationcolor]{0.13cm}{0.13cm} \textbf{vegetation} (41.99\%)  &\textcolor{blue}{\textbf{45.68}} &35.16& 	28.99 &	42.98& 	32.75& 	26.15&	44.85 &\textcolor{blue}{\textbf{39.82}}&\textcolor{blue}{\textbf{29.61}} \\

\crule[terraincolor]{0.13cm}{0.13cm} \textbf{terrain} (7.10\%) &24.70 	 &18.53 	 &14.69  &\textcolor{blue}{\textbf{31.96}}  &21.63  &16.75 &	25.08 	 &\textcolor{blue}{\textbf{22.13}}  &\textcolor{blue}{\textbf{21.59}} \\

\crule[polecolor]{0.13cm}{0.13cm} \textbf{pole} (0.22\%) & 8.84 &	8.16 	&\textcolor{blue}{\textbf{6.51}} &	9.28 &	8.45 &	6.92&\textcolor{blue}{\textbf{10.61}} &\textcolor{blue}{\textbf{9.15}} &	5.97 \\

\crule[trafficsigncolor]{0.13cm}{0.13cm} \textbf{traf.-sign} (0.06\%) &9.15 &\textcolor{blue}{\textbf{9.02}} &	6.92& 	8.58& 	7.67 &	5.67  &\textcolor{blue}{\textbf{12.98}}& 	8.32 	&\textcolor{blue}{\textbf{7.71}} \\

\crule[other-struct.color]{0.13cm}{0.13cm} \textbf{other-struct.} (4.33\%) & 10.31 &7.02 &	3.79& 	9.18& 	6.76 &	4.20  &\textcolor{blue}{\textbf{17.18}} 	&\textcolor{blue}{\textbf{8.97}} 	&\textcolor{blue}{\textbf{5.24}} \\

\crule[other-objectcolor]{0.13cm}{0.13cm} \textbf{other-obejct} (0.28\%) &4.40  & 3.27 &	2.43 	&\textcolor{blue}{\textbf{5.86}} 	&\textcolor{blue}{\textbf{4.49}} &	3.09&	5.53 &	3.75 	&\textcolor{blue}{\textbf{3.51}}  \\
\bottomrule
\end{tabular}
}
\vspace{-0.3cm}
\caption{\textbf{Quantitative comparison on SSCBench-KITTI-360~\cite{li2023sscbench}.} We present the results for various distance intervals (12.8 meters, 25.6 meters, and 51.2 meters) and furnish metrics for both geometric evaluation (IoU) and semantic assessment (mIoU). The best performance is highlighted in bold.}
\label{tab:quant-kitti360}
\vspace{-0.5cm}
\end{table}

\vspace{-0.6cm}
\vspace{-0.1cm}
\paragraph{SSCBench-KITTI-360} Analyzing the results for DepthSSC on the SSCBench-KITTI-360 dataset, as shown in Table~\ref{tab:quant-kitti360}, it is evident that DepthSSC excels particularly in recognizing roads and sidewalks. This demonstrates DepthSSC's strong capacity in interpreting and completing scenes involving larger, more contiguous structures. DepthSSC's comparative advantage over MonoScene in the 'Truck', 'Motorcycle', and 'Truck' categories is consistent at different ranges, illustrating its enhanced dynamic information processing capabilities. At longer ranges, DepthSSC's advantages over other benchmark models become more pronounced due to its ability to effectively utilize scene geometry information to correct inaccurate predictions at a distance.

\vspace{-0.1cm}

\subsection{Qualitative Visualizations}
\vspace{-0.1cm}
Figure~\ref{fig:vis} presents a visual comparison of scene completion by different models on the SemanticKITTI~\cite{behley2019semantickitti} validation dataset. In the first row, DepthSSC accurately predicts the car ahead, even under severe occlusion. Other models, however, show varying degrees of missed or misidentified vehicles, often confusing them with trees or similar objects. In the second column, VoxFormer provides inaccurate shape predictions for the nearby car, predicting only a few grid cells. In contrast, DepthSSC delivers highly accurate and continuous shape predictions. In the third row, DepthSSC is the only model that correctly identifies a pedestrian on the right side, partially obstructed by a car, demonstrating its ability to fully utilize surrounding scene information.

\subsection{Ablation Studies}
\label{ablation}

\begin{table}[t]
    \centering
    \footnotesize
    \resizebox{\columnwidth}{!}{
    \begin{tabular}{l|c|cc|cc}
        \toprule
        & \textbf{Experiment Type} & \textbf{Params (M)} & \textbf{FLOPs (G)} &   \multicolumn{2}{c}{\textbf{SemanticKITTI}} \\
        \textbf{Methods} & (Ablation/Alternative) &  &  & IoU $\uparrow$ & mIoU $\uparrow$ \\
        \midrule
        \textbf{Ours} & Full Model & \textbf{85.46} & \textbf{628.34} & \textbf{45.97}  & \textbf{14.59} \\

        \textbf{w/o Dynamic Resolution} & Ablation & 80.12  & 600.45  & 45.43 \textcolor{blue}{(-0.54)} & 13.92 \textcolor{blue}{(-0.67)} \\

        \textbf{w/o GAV}   & Ablation & 80.05 & 610.22  & 45.30 \textcolor{blue}{(-0.67)} & 13.74 \textcolor{blue}{(-0.85)} \\

        \textbf{w/o ST-GF} & Ablation & 63.14  & 540.18  & 45.14 \textcolor{blue}{(-0.83)} & 13.55 \textcolor{blue}{(-1.04)} \\

        \midrule

        \textbf{Position Offset Prediction (POP)} & Alternative & 61.23 & 535.70 & 44.85 \textcolor{blue}{(-1.12)} & 13.10 \textcolor{blue}{(-1.49)} \\

        \textbf{Depth-Aware Warping (DAW)} & Alternative & 62.31  & 533.48  & 44.62 \textcolor{blue}{(-1.35)} & 12.97 \textcolor{blue}{(-1.62)} \\

        \textbf{Deformable Voxel Alignment (DVA)} & Alternative & 62.05 & 532.90  & 44.50 \textcolor{blue}{(-1.47)} & 12.78 \textcolor{blue}{(-1.81)} \\

        \textbf{VoxFormer}~\cite{li2023voxformer} & Baseline & 57.81 & 530.24  & 44.02 \textcolor{blue}{(-1.95)} & 12.35 \textcolor{blue}{(-2.24)} \\

        \bottomrule
    \end{tabular}}
    \vspace{-0.3cm}
    \caption{\textbf{Ablation and alternative methods evaluation.} The table compares the performance of ablation studies and alternative methods on SemanticKITTI~\cite{behley2019semantickitti}. ST-GF shows stronger performance than simpler alternatives like POP, DAW, and DVA, highlighting its advantage in capturing spatial dependencies. The addition of the dynamic resolution in GAV also contributes significantly to the final performance.}
    \vspace{-0.8cm}
    \label{tab:ablation_com}
\end{table}

To validate DepthSSC, we conducted ablation studies. Table~\ref{tab:ablation_com} demonstrates significant performance improvements from both modules. The ST-GF module enhances IoU and mIoU by 0.83 and 1.04, respectively, by ensuring accurate spatial alignment between the depth map and voxel queries through spatial transformation and graph structure optimization. The GAV module, by dynamically adjusting voxel resolution based on geometric complexity, improves IoU and mIoU by 0.67 and 0.85, respectively. This adaptive resolution captures more details in complex regions while conserving resources in simpler areas, enhancing VoxFormer's overall modeling capability. For more detailed ablation studies on the design of the DepthSSC model, please refer to the supplementary materials. Additionally, we can observe that the Params (M) increased with the inclusion of ST-GF, while GAV only caused a minor increase in parameters. However, the FLOPs (G) slightly decreased with GAV, as it saves computation by adjusting voxel sizes dynamically. The overall FLOPs for the full model are acceptable.

We experimented with three alternative approaches to ST-GF: (1) \textbf{Position Offset Prediction (POP)}, which directly predicts voxel offsets based on 3D position differences from the input depth; (2) \textbf{Depth-Aware Warping (DAW)}, which warps voxel positions by reprojecting them into the 2D image plane, adjusting the 3D coordinates based on estimated depth differences between the projected and original voxels; (3) \textbf{Deformable Voxel Alignment (DVA)}, which locally applies deformable transformations to adjust voxel positions based on local geometric features and depth cues. The results show that although these alternatives exhibit some effectiveness, their performance is inferior compared to ST-GF. 

To further analyze the contribution of dynamic resolution \(R(V)\) introduced by the GAV module, we compared the performance of DepthSSC without the dynamic resolution (fixed resolution across all voxels). Table~\ref{tab:ablation_com} shows that removing dynamic resolution causes a drop in IoU by 0.54 and mIoU by 0.67. This confirms that the dynamic resolution component in GAV significantly contributes to capturing intricate geometries.
\vspace{-0.2cm}
\subsection{Robustness experiment}
\vspace{-0.1cm}
To evaluate the robustness of the ST-GF module under depth input errors, we simulate errors in depth measurements by artificially introducing Gaussian noise with varying intensities into the depth input, defined as $N(0, \sigma^2)$, where $\sigma$ represents different noise levels. We then observe the performance of the ST-GF module under different noise levels. The experimental results are shown in Figure \ref{tab:noise_robustness}. Despite adding different levels of noise to the depth input, the performance of the ST-GF module remains stable, with less than 1.0\% drop in mIoU under mild and moderate noise conditions, demonstrating strong robustness. In contrast, the alternative methods show a decline in mIoU of over 1.5\% under moderate and severe noise conditions. This robustness experiment verifies that the ST-GF module is better at capturing spatial dependencies between voxels and exhibits stronger resistance to depth noise.

\begin{table}[t]
	\centering
	\footnotesize
	\resizebox{\columnwidth}{!}{
	\begin{tabular}{l|c|c|c}
		\toprule
		\textbf{Noise Level ($\sigma$)} & \textbf{Methods} & \textbf{mIoU $\uparrow$} & \textbf{Degradation (mIoU) $\downarrow$} \\
		\midrule
		\textbf{No Noise} & Ours (ST-GF) & \textbf{14.59}  & -- \\
		\midrule
		\multirow{4}{*}{\scalebox{0.85}[1]{\( \sigma = 0.01 \)}} 
		& Ours (ST-GF) & 14.45 &  \textcolor{blue}{-0.14}  \\
		& Position Offset Prediction (POP) & 12.07 & \textcolor{blue}{-1.03} \\
		& Depth-Aware Warping (DAW) & 11.85 & \textcolor{blue}{-1.12} \\
		& Deformable Voxel Alignment (DVA) & 11.29 & \textcolor{blue}{-1.49} \\
		\midrule
		\multirow{4}{*}{\scalebox{0.85}[1]{\( \sigma = 0.05 \)}} 
		& Ours (ST-GF) & 14.12 & \textcolor{blue}{-0.47}  \\
		& Position Offset Prediction (POP) & 11.42 & \textcolor{blue}{-1.68} \\
		& Depth-Aware Warping (DAW) & 11.30 & \textcolor{blue}{-1.67} \\
		& Deformable Voxel Alignment (DVA) & 10.53 & \textcolor{blue}{-2.25} \\
		\midrule
		\multirow{4}{*}{\scalebox{0.85}[1]{\( \sigma = 0.1 \)}} 
		& Ours (ST-GF) & 13.76 & \textcolor{blue}{-0.83}  \\
		& Position Offset Prediction (POP) & 10.96 & \textcolor{blue}{-2.14} \\
		& Depth-Aware Warping (DAW) & 10.64 & \textcolor{blue}{-2.33} \\
		& Deformable Voxel Alignment (DVA) & 10.26 & \textcolor{blue}{-2.52} \\
		\bottomrule
	\end{tabular}}
 \vspace{-0.2cm}
 \caption{\textbf{Robustness evaluation under noisy depth inputs.} This table shows the performance degradation in mIoU under increasing depth noise levels. Our ST-GF module demonstrates higher robustness compared to simpler alternatives.}
\vspace{-0.6cm}
	\label{tab:noise_robustness}
\end{table}

\section{Conclusion}

In this study, we presented DepthSSC for monocular semantic scene completion that leverages spatial transformation and feature fusion to accurately align spatial and depth information. By integrating the Spatially-Transformed Graph Fusion (ST-GF) module and Geometrically-aware Voxelization, DepthSSC dynamically adjusts voxel resolutions based on the geometric complexity of 3D space, addressing the limitations of existing methods like VoxFormer, such as spatial misalignment, object boundary distortion, and inadequate depth perception. Our approach effectively mitigates these issues, enhancing the accuracy of 3D scene reconstruction. Rigorous evaluations on the SemanticKITTI and SSCBench-KITTI-360 dataset demonstrate that DepthSSC captures intricate 3D structural details and achieves state-of-the-art performance. We believe that DepthSSC will inspire further research and advancements in monocular camera-based semantic scene completion.

%
%

{\small
\bibliographystyle{ieee_fullname}
\bibliography{main}
}

\clearpage

\appendix

\section{Experiment setup}

\label{dataset}
\paragraph{Dataset} We evaluate DepthSSC on the SemanticKITTI~\cite{behley2019semantickitti} dataset, renowned for its dense semantic annotations of urban driving sequences from the KITTI Odometry Benchmark. The dataset voxelizes point clouds into a 51.2m×51.2m×64m scene, represented by 256×256×32 voxel grids, and includes 20 semantic classes, including the "empty" category. SemanticKITTI includes RGB images (1220×370) and LiDAR sweeps as inputs. The dataset is divided into 10 sequences for training, 1 for validation, and 11 for testing. We also evaluate DepthSSC on the SSCBench-KITTI360\cite{li2023sscbench},which consist of 20 and 19 classes. This dataset consist of voxel grids with semantic labels covering an area of $51.2 \, \text{m} \times 51.2 \, \text{m} \times 6.4 \, \text{m}$, with each voxel having a size of $0.2 \, \text{m}$, creating a grid resolution of $256 \times 256 \times 32$.

\paragraph{Metric} For our experimentation with the DepthSSC model, we exclusively utilize RGB images from a monocular vision setup. These images, being a primary source of input for our model, facilitate the understanding of scene structures and semantics. Our primary evaluation metrics remain focused on the intersection over union (IoU) for the occupied voxel grids. Additionally, we also adopt the mean IoU (mIoU) metric for voxel-wise semantic evaluations.

\paragraph{Baselines} We compare our proposed DepthSSC with existing SSC baselines (JS3CNet~\cite{roldao2020lmscnet}, AICNet~\cite{li2020anisotropic} and LMSCNet~\cite{xu2021sparse}). We also compare DepthSSC with MonoScene~\cite{cao2022monoscene}, TPVFormer~\cite{li2020anisotropic}, VoxFormer~\cite{li2023voxformer}, NDC-Scene~\cite{yao2023ndc}, Symphonies\cite{jiang2024symphonize} and HASSC\cite{wang2024not}, which are best RGB-only SSC methods. Note that for the methods with more than RGB inputs, we follow~\cite{cao2022monoscene} to adapt their results to RGB only inputs. 

Moreover, in Table 1 and Table 2 of the main paper, the notations Occ, Depth and Pts denote the occupancy grid, depth map and point cloud, which are the 3D input required by the SSC baselines. For a fair comparison, all the three inputs are converted from the depth map predicted by a pretrained depth predictor~\cite{bhat2021adabins}.For implementation details of DepthSSC, please refer to the supplementary materials.

\begin{table}[t]
	\centering
        \captionsetup{font=footnotesize}
	\setlength{\tabcolsep}{0.010\linewidth}
	\centering
	\footnotesize

	\begin{tabular}{l|cc}
		\toprule
		 & \multicolumn{2}{c}{Ours(SemanticKITTI)} \\
		Alignment Method & IoU $\uparrow$ & mIoU $\uparrow$ \\
		\midrule
			
	ST-GF   & \textbf{45.97}  & \textbf{14.59} 	
		\\
		
		ICP & 44.28 \textcolor{blue}{(-1.69)} & 12.64\textcolor{blue}{(-1.95)}
		\\
		Feature-based Reg.  & 44.45\textcolor{blue}{(-1.52)} & 12.76\textcolor{blue}{(-1.83)}
		\\
		Regularization Matching  &  44.87\textcolor{blue}{(-1.10)} & 12.98\textcolor{blue}{(-1.61)}	
		\\
		
		\bottomrule
	\end{tabular}
	 \caption{\textbf{Ablation study} for ASAN in ST-GF module.}

	\label{tab:ablation_STN}
\end{table}

\begin{table}[ht]
	\centering
        \captionsetup{font=footnotesize}
	\setlength{\tabcolsep}{0.010\linewidth}
	\centering
	\footnotesize
 
	\begin{tabular}{l|cc}
		\toprule
		 & \multicolumn{2}{c}{Ours(SemanticKITTI)} \\
		Distance Metric & IoU $\uparrow$ & mIoU $\uparrow$ \\
		\midrule
			
	Euclidean Distance   & \textbf{45.97}  & \textbf{14.59} 	
		\\
		
		Cosine Similarity & 45.46\textcolor{blue}{(-0.51)} & 13.25\textcolor{blue}{(-1.34)}
		\\
		Manhattan Distance  & 45.33\textcolor{blue}{(-0.64)} & 12.92\textcolor{blue}{(-1.67)}
		\\
		
		\bottomrule
	\end{tabular}
	\caption{\textbf{Ablation study} on connection strength computation using different distance metrics.}

	\label{tab:ablation_dm}
\end{table}

\begin{table*}[t]
\centering
\renewcommand{\arraystretch}{0.9}  
\setlength{\tabcolsep}{1.5pt}  
\scriptsize  
\begin{tabular}{l|ccccccccccccccccc|c}
\hline
Method  & \rotatebox{90}{\textcolor{black}{$\blacksquare$} others} & \rotatebox{90}{\textcolor[RGB]{255,120,50}{$\blacksquare$} barrier} & \rotatebox{90}{\textcolor[RGB]{255,192,203}{$\blacksquare$} bicycle} & \rotatebox{90}{\textcolor[RGB]{255,255,0}{$\blacksquare$} bus} & \rotatebox{90}{\textcolor[RGB]{0,150,245}{$\blacksquare$} car} & \rotatebox{90}{\textcolor[RGB]{0,255,255}{$\blacksquare$} const. veh.} & \rotatebox{90}{\textcolor[RGB]{255,127,0}{$\blacksquare$} motorcycle} & \rotatebox{90}{\textcolor[RGB]{255,0,0}{$\blacksquare$} pedestrian} & \rotatebox{90}{\textcolor[RGB]{255,240,150}{$\blacksquare$} traffic cone } & \rotatebox{90}{\textcolor[RGB]{135,60,0}{$\blacksquare$} trailer} & \rotatebox{90}{\textcolor[RGB]{160,32,240}{$\blacksquare$} truck} & \rotatebox{90}{\textcolor[RGB]{255,0,255}{$\blacksquare$} drive. surf.} & \rotatebox{90}{\textcolor[RGB]{139,137,137}{$\blacksquare$} other flat} & \rotatebox{90}{\textcolor[RGB]{75,0,75}{$\blacksquare$} sidewalk} & \rotatebox{90}{\textcolor[RGB]{150,240,80}{$\blacksquare$} terrain} & \rotatebox{90}{\textcolor[RGB]{230,230,250}{$\blacksquare$} manmade} & \rotatebox{90}{\textcolor[RGB]{0,175,0}{$\blacksquare$} vegetation} & \rotatebox{90}{mIoU} \\ \hline
MonoScene\cite{cao2022monoscene} &1.75&7.23&4.26&4.93&9.38&5.67&3.98&3.01&5.90&4.45&7.17&14.91&6.32&7.92&7.43&1.01&7.65&6.06\\
TPVFormer\cite{huang2023tri}  &7.22&38.90&13.67&\textbf{40.78}&45.90&17.23&19.99&18.85&14.30&26.69&34.17&55.65&35.47&37.55&30.70&19.40&16.78&27.83 \\
BEVDet\cite{huang2021bevdet}  &4.39&30.31&0.23&32.36&34.47&12.97&10.34&10.36&6.26&8.93&23.65&52.27&24.61&26.06&22.31&15.04&15.10&19.38 \\
OccFormer\cite{zhang2023occformer}  &5.94&30.29&12.32&34.40&39.17&14.44&16.45&17.22&9.27&13.90&26.36&50.99&30.96&34.66&22.73&6.76&6.97&21.93 \\
BEVFormer\cite{li2022bevformer}  &5.85&37.83&17.87&40.44&42.43&7.36&23.88&21.81&20.98&22.38&30.70&55.35&28.36&36.0&28.06&20.04&17.69&26.88 \\
DepthSSC(Ours) &\textbf{12.11}&\textbf{40.26}&\textbf{28.35}&37.42&\textbf{54.56}&\textbf{23.15}&\textbf{29.20}&\textbf{28.73}&\textbf{29.21}&\textbf{31.22}&\textbf{38.67}&\textbf{72.88}&\textbf{46.05}&\textbf{47.85}&\textbf{33.23}&\textbf{37.02}&24.37&\textbf{38.84}      \\ \hline
\end{tabular}
\caption{3D semantic occupancy prediction performance on the validation set of Occ3D-nuScenes\cite{tianOcc3DLargeScale3D2023}.}
\label{sota_comparison}
\end{table*}
\section{Implementation details}
\subsection{Architectures}
We adopt ResNet-50~\cite{he2016deep} as the backbone for 2D feature extraction. The backbone consists of four stages, and we utilize features from the third stage (out\_indices=(2,)) for further processing. The network is partially frozen (frozen\_stages=1) and employs batch normalization (BN) for stable training. We employ a Feature Pyramid Network (FPN)~\cite{lin2017feature} to process the extracted 2D features. The FPN takes the 1024-dimensional features from the ResNet backbone and transforms them into a 128-dimensional feature space. The FPN starts from level 0, adds extra convolutions on output, and produces feature maps with a spatial resolution of \( H/4 \times W/4 \times 128 \).

In the proposed method, voxel queries are 3D grid-shaped learnable parameters that map 2D features to the 3D volume. The voxel queries \( Q \in \mathbb{R}^{64 \times 64 \times 16 \times 128} \) are generated at a lower resolution to reduce computational load. From these voxel queries, a subset \( Q_p = \text{Reshape}(Q[M_{\text{out}}]) \) is selected based on predicted occupancy from depth information, resulting in \( Q_p \in \mathbb{R}^{1024 \times 128} \), where \( M_{\text{out}} \) is the corrected occupancy map. To handle multi-modal data, we incorporate a cross-transformer and a self-transformer. The cross-transformer utilizes a PerceptionTransformer architecture with three encoder layers, each based on VoxFormerEncoder, which processes input using deformable cross-attention mechanisms~\cite{kim2023cross} to integrate multi-view image features into a unified 3D space. This encoder attends to 8 points per pillar and employs Multi-Scale Deformable Attention~\cite{zhu2020deformable} for effective feature fusion. Each layer in the cross-transformer has an embedding dimension of 128 and a feedforward dimension of 256, with a dropout rate of 0.1. The self-transformer follows a similar PerceptionTransformer3D architecture with two encoder layers, using deformable self-attention to refine voxel features within the 3D space. The self-transformer has an embedding dimension of 128 and attends to 8 points within each voxel.

Our Spatially-Transformed Graph Fusion (ST-GF) module addresses the misalignment issue between depth maps and voxel queries. The Adaptive Spatial Adjustment Network (ASAN) predicts a 3D affine transformation matrix \( \Theta \) for each voxel query \( Q \in \mathbb{R}^{64 \times 64 \times 16 \times 128} \) and depth prediction \( D \in \mathbb{R}^{64 \times 64 \times 16 \times 1} \). Using \( \Theta \), the grid generator maps the output space to the input space, applying trilinear interpolation to adjust voxel positions. Transformed voxels are clustered into nodes, with edges representing spatial relationships. Node features are fused via graph convolution, and the refined features are backpropagated to the original voxel space, ensuring accurate scene comprehension. The resolution-adaptive deformable attention mechanism adjusts the positions and quantities of query points in deformable self-attention based on the dynamically assigned resolution. For each voxel \( V_i \), its position in three-dimensional space can be represented as \( p = (x, y, z) \). We adjust these positions based on the resolution \( R(V_i) \), allowing voxels with higher complexity to have a higher query density. The adjusted query points are calculated as \( p' = p + \delta R(V_i) + \Delta p \), where \( \delta = 0.1 \) is a constant.

\subsection{Hyperparameter for Training}
We utilize 4 NVIDIA Tesla A100 GPUs to train the DepthSSC model across 30 epochs, processing a batch size of 4 images in each iteration. These RGB images are of the resolution 1220×370. During training, we incorporate a random horizontal flip for data augmentation. For optimization, we employ the AdamW optimizer, initiating with a learning rate of 1e-4 coupled with a weight decay of 1e-4. By the time we reach the 5th epoch, we decrease the learning rate by 10\%. Both stage-1 and stage-2 are trained separately for 24 epochs, using a learning rate of $2 \times 10^{-4}$.

\section{Additional Ablation Studies}

\paragraph{ST-GF Ablation Experiments.} The ST-GF module combines spatial transformation and graph structure features to ensure accurate alignment of spatial information between the depth map and voxel queries in 3D scene completion. Alternative alignment techniques, such as Iterative Closest Point (ICP)\cite{zhang2021fast}, feature-based registration\cite{kuppala2020overview}, and regularization-based matching\cite{liu2015regularization}, can also be used. Regularization-based matching minimizes a distance metric between the source and target, while feature-based registration uses extracted feature points for matching. As shown in Table~\ref{tab:ablation_STN}, ST-GF outperforms other alignment methods, demonstrating its effectiveness. However, any alignment method improves the performance of the original VoxFormer.

\paragraph{Connection Strength Ablation Experiments.} In the ST-GF module, connection strength represents relationships or similarities between nodes, influencing which relationships are fused during graph convolution. We compare different distance metrics for semantic scene completion (SSC). Table~\ref{tab:ablation_dm} shows that Euclidean distance achieves the best performance, as it accurately captures the actual distance and relative positional relationships in 3D space. Cosine similarity, focusing more on direction than magnitude, is less suitable for this task. Manhattan distance considers spatial aspects but does not account for the shortest distance between two points, leading to potential information loss or inaccuracies.

\begin{table}[t]
	\centering
        \captionsetup{font=footnotesize}
	\setlength{\tabcolsep}{0.010\linewidth}
	\centering
	\footnotesize

	\begin{tabular}{l|cc}
		\toprule
		 & \multicolumn{2}{c}{Ours(SemanticKITTI)} \\
		Region-Adaptive Method & IoU $\uparrow$ & mIoU $\uparrow$ \\
		\midrule
			
	Resolution-Adaptive Deformable Attention   & \textbf{45.97}  & \textbf{14.59} 	
		\\
		
		Non-Uniform Voxelization & 40.61\textcolor{blue}{(-5.36)} & 10.63\textcolor{blue}{(-3.96)}
		\\
		Dynamic Kernel Methods  & 42.58\textcolor{blue}{(-3.39)} & 11.80\textcolor{blue}{(-2.79)}
		\\
        Non-Local Operations  & 41.47\textcolor{blue}{(-4.50)} & 11.39\textcolor{blue}{(-3.20)}
		\\
		
		\bottomrule
	\end{tabular}
	
 \caption{\textbf{Ablation study} on resolution-adaptive deformable attention.}
	\label{tab:ablation_da}
\end{table}

\paragraph{Resolution-Adaptive Deformable Attention Ablation Experiments.} Resolution-adaptive deformable attention addresses the varying geometric complexities in 3D data by enabling finer voxel resolutions in complex regions. To validate its effectiveness, we compared it against non-uniform voxelization\cite{hu2022voxel}, dynamic kernel\cite{cheng2021s3cnet}, and non-local operations\cite{park2020non}. These methods partially address geometric complexity but have limitations. Table~\ref{tab:ablation_da} presents the ablation results. Non-uniform voxelization can cause data discontinuities and biases. Dynamic kernel methods face alignment issues with different kernel sizes and shapes. Non-local operations, while capturing long-range dependencies, are computationally intensive for 3D data and less effective at local complexities. The results demonstrate that resolution-adaptive deformable attention outperforms these methods, capturing local geometric details more effectively.

\section{Results on Occ3D} The results in Table~\ref{sota_comparison} demonstrate the superiority of our DepthSSC over several state-of-the-art methods on the Occ3D-nuScenes validation set. DepthSSC achieves the highest mIoU of 38.84, outperforming existing approaches such as TPVFormer (27.83) and BEVFormer (26.88) by a significant margin.

Specifically, ST-GF allows for more accurate spatial alignment between voxel queries and depth maps, leading to enhanced object recognition in challenging categories such as vegetation (37.02) and manmade structures (33.23). GAV further refines voxel resolution dynamically, which is particularly beneficial in capturing fine details in complex categories like bicycles (28.35) and pedestrians (28.73). DepthSSC also excels in driveable surface detection (72.88), indicating that the proposed fusion and voxelization techniques are effective in both object-level and scene-level predictions. The substantial improvements across a range of categories, especially in highly dynamic or small-scale objects, validate the robustness of our approach in semantic scene completion tasks.

\end{document}